\documentclass[11pt]{asaproc}

\usepackage[ruled,vlined]{algorithm2e}

\usepackage{graphics}
\usepackage{graphicx}
\usepackage{grffile}

\usepackage[utf8]{inputenc}
\usepackage[english]{babel}
\usepackage[backend=bibtex]{biblatex}
\usepackage{makecell}
\usepackage{csquotes}
\usepackage{tikz}
\usepackage[latex]{xellipsis}
\usepackage{multicol}
\usepackage{amsmath}
\usepackage{floatrow}
\usepackage{mathtools} 
\usepackage{multirow}
\usepackage{array}
\usepackage{float}
\usepackage{subcaption}
\usepackage{amsthm}
\graphicspath{ {figures/} }
\usepackage{comment}
\usepackage{hyperref}

\newcommand{\nth}[1]{#1^{\text{th}}}
\newcommand{\splitatcommas}[1]{\begingroup\lccode`~=`, \lowercase{\endgroup
    \edef~{\mathchar\the\mathcode`, \penalty0 \noexpand\hspace{0pt plus 1em}}%
  }\mathcode`,="8000 #1%
  }

\title{Density-based interpretable hypercube region partitioning for mixed numeric and categorical data}

\author{Samuel Ackerman\thanks{IBM Research, Haifa; samuel.ackerman@ibm.com} \and  Eitan Farchi\thanks{IBM Research, Haifa} \and  Orna Raz\footnotemark[2] \and Marcel Zalmanovici\footnotemark[2] \and Maya Zohar \footnotemark[2]}
\begin{document}

\maketitle

\begin{abstract}
Consider a structured dataset of features, such as $\{\textrm{SEX}, \textrm{INCOME}, \textrm{RACE}, \textrm{EXPERIENCE}\}$.  A user may want to know where in the feature space observations are concentrated, and where it is sparse or empty. The existence of large sparse or empty regions can provide domain knowledge of soft or hard feature constraints (e.g., what is the typical income range, or that it may be unlikely to have a high income with few years of work experience).  Also, these can suggest to the user that machine learning (ML) model predictions for data inputs in sparse or empty regions may be unreliable.

An interpretable region is a hyper-rectangle, such as $\{\textrm{RACE} \in\{\textrm{Black}, \textrm{White}\}\}\:\&$   \\$\{10 \leq \:\textrm{EXPERIENCE} \:\leq 13\}$, containing all observations satisfying the constraints; typically, such regions are defined by a small number of features.  Our method constructs an observation density-based partition of the observed feature space in the dataset into such regions. It has a number of advantages over others in that it works on features of mixed type (numeric or categorical) in the original domain, and can separate out empty regions as well.

As can be seen from visualizations, the resulting partitions accord with spatial groupings that a human eye might identify; the results should thus extend to higher dimensions.  We also show some applications of the partition to other data analysis tasks, such as inferring about ML model error, measuring high-dimensional density variability, and causal inference for treatment effect.  Many of these applications are made possible by the hyper-rectangular form of the partition regions.

\end{abstract}

\section{Introduction
\label{sec:intro}}

The task of conducting data exploration, visualization, and analysis, on a structured feature dataset can be challenging. The difficulties often increase when the data is high-dimensional and has mixed feature types (numeric, nominal categorical, ordinal categorical).  Furthermore, it is often difficult to provide analyses that are in a meaningful format for a human user to interpret and gain insights into the data.

Here, we outline an algorithm that describes a dataset of arbitrary feature dimension size, where the features may be of mixed type, by partitioning the feature space into regions according to the relative density of observed points.  Such regions will be presented in a form that is intuitive for human interpretation.  From these regions, a person can understand where in the potential space most of the data are.  Our method also finds empty regions of the same form, where data records are not observed to exist; these may be empty due to domain-specific feature constraints (i.e., a knowledgeable person would expect them to be empty), but they may still be of interest to the user.

The paper proceeds as follows: Section~\ref{sec:problem} introduces the relevant notation for describing feature distributions, constructing the hyper-rectangular regions, and calculations such as length and volume that can be performed on them.  The ability to do these calculations, and their intuitive nature, motivates much of the rest of the work.  Section~\ref{sec:algorithm} outlines the elements of the algorithm, in calculating observation density, and partitioning into empty and non-empty regions.  Section~\ref{sec:alternatives} compares our method's properties with those of some existing ones. Section~\ref{sec:examples} illustrates the partition method on a two-dimensional example, with varying parameter settings.  Section~\ref{sec:metrics} gives an example of a measurement of density variability that can be performed on a partition of regions, which can capture complex characteristics of the dataset.  Section~\ref{sec:applications} gives two examples of applications of the density partition to other data analysis tasks: inferring ML performance on unlabeled data, and performing causal analysis on treatment outcomes.  The Appendix gives more detail about region length and volume calculations, and contains an example of a partition with these calculations worked out step-by-step.
\section{Problem description
\label{sec:problem}}

Here we outline the relevant mathematical notation to illustrate our method.  Let $D$ be a tabular dataset of $p$ features, $\{F_1,\dots,F_p\}$ and $n$ observations $\mathbf{x}_i,\:i=1,\dots,n$, each of dimension $p$.
Our method will return a set of regions, or `slices', defined on the dataset feature, where regions are separated by varying levels of observation density.  Some regions can be empty, and together they partition, or divide between them, the feature space as contained in the dataset $D$.  These regions will be human-interpretable and in the original feature domains, without having to distort or grid the data.  We thus have to define concepts such as the features' domains (Section~\ref{ssec:feature_domains}), building slices from subsets of individual features (Section~\ref{ssec:domain_subsets}), and the size of subsets and slice volumes (Section~\ref{ssec:slice_size}).

The $n$ observations in $D$ are assumed to be mutually independent and their row indices interchangeable; the method outlined will not deal with specific cases such as time or spatial indices of observations, such as image pixel data coded in matrix form, where pixels (feature values in a given row)  have a spatial dependency to neighboring pixels. 

\subsection{Feature domains
\label{ssec:feature_domains}}

Our method accommodates four data feature types, with differential treatment of each, as we will see:
numeric features can be either continuous real-valued or integer-valued, and categorical features can be either nominal (no order) or ordered.  The user should specify if integer-valued features are actually numeric or are codes for categorical levels.
Let $\textrm{dom}(F_j),\: j=1,\dots,p$ be the (marginal) domain of feature $F_j$, as observed in the dataset $D$.  It is assumed that $\textrm{dom}(F_j)$ represents the full range of potential values of the feature $F_j$; that is, that the range as given by the sample observed in $D$ is the range in the hypothetical population distribution of $F_j$.     $\textrm{dom}(F_j)$ is defined as

\begin{itemize}
    \item $[\textrm{min}(F_j),\:\textrm{max}(F_j)]$, where $\textrm{min}(F_j)$ and $\textrm{max}(F_j)$ are the minimum and maximum observed values of $F_j$ in $D$, if $F_j$ is \textbf{real-valued}.  If $F_j$ is strictly numeric \textbf{integer-valued}, this is modified slightly to $[\textrm{min}(F_j)-0.5,\:\textrm{max}(F_j)+0.5]$.
    \item $\{\textrm{unique}(F_j)\}$, the set of unique observed values in $D$, if $F_j$ is \textbf{nominal categorical}.
    \item $F_j$ may be \textbf{ordered categorical} with an ordered set of levels $\boldsymbol{\ell}=\{\ell_1,\dots,\ell_K\}$ (user input). Let $\textrm{min}(F_j)=\ell_a$ and  $\textrm{max}(F_j)=\ell_b$, where $1\leq a<b \leq K$.  The values $\{\ell_a,\dots,\ell_b\}$ are recoded to integers $\{0,\dots,b-a\}$, so the domain is $[-0.5, (b-a)+0.5]$.  Note that because the full set of levels was pre-specified and that the levels have order, $\textrm{dom}(F_j)$ may include intermediate values not observed in $D$, unlike the nominal categorical domain, where only observed values are included.  Thus, the treatment of ordered categorical is similar to integers.
\end{itemize}

We assume the feature values are either fully-ordered (numeric or ordered categorical) or not ordered (nominal categorical).  We do not attempt to deal with partial orderings, though, in some cases these could be treated as categorical with some values being full ordered and others being nominal.  This is because, as discussed below, we need to know which values are contained in a given domain subset.

\subsection{Feature domain subsets, slices, and the potential feature space
\label{ssec:domain_subsets}}

Let $s_{i,j} \subseteq\textrm{dom}(F_j)$ be a given  properly-defined subset, or interval, of a given feature domain; $s_{i,j}$ can trivially be the full domain $\textrm{dom}(F_j)$ itself.  Below, we will consider several different intervals, indexed by $i$, on the same feature $F_j$.  For instance, if $F_1$ ($j=1$) is INCOME, with $\textrm{dom}(F_1)=[\$0,\: \$500{,}000]$, a legal subset could be $s_{1,1}=[\$0,\: \$30{,}000]$, containing relatively low-income individuals; for feature $F_3$=STATE with $\textrm{dom}(F_3)=\{\splitatcommas{\textrm{Alabama},\dots,\textrm{Wyoming}}\}$, one possibility is $s_{1,3}=\{\textrm{California}, \:\textrm{New York}\}$.   

A hyper-rectangle, or `slice' $S_i$ is defined as the mutual intersection of a set of $\{s_{i,j}\}_{j=1}^p$, one on each feature $F_j$; that is,
\[S_i\equiv\bigcap_{j=1}^p s_{i,j} =s_{i,1}\bigcap\dots\bigcap s_{i,p}\]
for a particular choice of subsets.  If $s_{i,j}=\textrm{dom}(F_j)$ for one or more features $j$, these subsets are `complete', and these features are considered to not define $S_i$ since there is no restriction on their values within $\textrm{dom}(F_j)$.  The number of non-complete subsets, that is $\sum_{j=1}^pI(s_{i,j}\ne \textrm{dom}(F_j))$, where $I(\cdot)$ denotes the indicator function, is the slice $S_i$'s dimension.  Any observation $\mathbf{x}=\begin{bmatrix}x_1 & \dots & x_p\end{bmatrix}$, whether from $D$ or from another similar dataset $D'$ with the same features, for which all feature subsets, or constraints, are satisfied, is said to belong to the slice.  Formally, $\mathbf{x}\in S_i$ iff $\left(\prod_{j=1}^p I(x_j\in s_{i,j})\right)=1$.  

For instance, the slice $S_1=\{\$0\leq \:\textrm{INCOME}\:\leq \$30{,}000\}\: \&\: \{\textrm{STATE}\in\{\splitatcommas{\textrm{California},\: \textrm{New York}}\}\}$ is formally defined as $\bigcap_{j=1}^p s_{1,j}$, where $s_{1,1}$ and $s_{1,3}$ on INCOME and STATE are as defined above, and $s_{1,j}=\textrm{dom}(F_j)$ for $j\notin\{1,3\}$ for the other features.  This slice contains all Californians and  New Yorkers with the given income range (that is, the intersection of these two subsets on the two features), regardless of the values of other features; this slice is of dimension 2.  For $\mathbf{x}=\begin{bmatrix}\splitatcommas{\$25{,}000, & 21, & \textrm{Colorado}, & \textrm{Student}, & \textrm{Female}}\end{bmatrix}$, $\mathbf{x}\notin S_1$ because while $x_1=\$25{,}000\in s_{1,1}$, $x_3=\textrm{Colorado}\notin s_{1,3}$. The observation is not in the slice if at least one of the feature constraints (here, the state), is not satisfied.  For the other features, $x_j\in s_{1,j}$ automatically because the slice does not use these features, and thus $s_{1,j}=\textrm{dom}(F_j)$, and by definition, any observed value for feature $F_j$ in $D$ is in the domain.

This specification of slices is motivated by our related work (e.g., \cite{ARZ2020}), which uses this definition to define data subsets with high rates of machine learning classifier error, which are returned to a user to help diagnose the classifier's areas (in terms of feature value combinations) of weakness.  There, the focus was on making the output `interpretable' to the user.  The logic is that a user can easily comprehend the meaning of slices such defined, particularly when the slice dimension is relatively low, say, no more than some $p^*$, such as 3 or 4; the same intuition applies here.

The potential feature space, as given by the observed $D$, is defined as $\mathcal{S}=\bigcap_{j=1}^p\textrm{dom}(F_j)$.  That is, $\mathcal{S}$ is the hyper-rectangle that inscribes, or is the tightest rectangular bound, on $D$; this is analogous to a rectangle that tightly bounds a set of points in 2-D Euclidean space, without allowing axis rotations.\footnote{This definition of $\mathcal{S}$ makes it sensitive to extreme outliers, since it is defined by the bounding observations in each direction.  In Section~\ref{ssec:density_proxy} we show how $\mathcal{S}$  
can be made more robust by re-defining it after pre-filtering the most extreme outliers.}

\subsection{Subset sizes and slice volumes
\label{ssec:slice_size}}

For each domain type, an appropriate size metric $|\textrm{dom}(F_j)|$ can be defined.  For integer and ordered categorical features, where $\textrm{dom}(F_j)=[a,\:b]$, $|\textrm{dom}(F_j)|:= b-a+1$; this corresponds to the unique potential number of integer or category values.  For continuous real-valued domains of form $\textrm{dom}(F_j)=[a,\:b]$, $|\textrm{dom}(F_j)|:= b-a$, the size of the interval on the real line.  For nominal categorical, $|\textrm{dom}(F_j)|$ is the length of the defining set, the number of unique categories observed.  Similarly, we can define the size $|s_{i,j}|$ of a proper subset $s_{i,j}\subset \textrm{dom}(F_j)$.  For integer, and nominal and ordered categorical features, $|s_{i,j}|$ is calculated in the same way as $|\textrm{dom}(F_j)|$.  For $s_{i,j}=[a,\:b]$, if $F_j$ is ordered categorical, this interval represents the range of categories between levels $a$ and $b$, some of which may not be observed; for integer-valued features $F_j$, $[a,\:b]$ contains all values between the integers $a+0.5$ and $b-0.5$.  For real-valued features, see Appendix~\ref{ssec:edge_length}.

So that measurements are standardized across features and datasets, the fractional length of a subset $s_{i,j}$ is defined as $L(s_{i,j})=|s_{i,j}|/|\textrm{dom}(F_j)|$, where $0<L(s_{i,j})\leq 1$, since $s_{i,j}$ must always be nonempty.  By definition, $L(\textrm{dom}(F_j))=1$.  Just as a rectangular prism has volume equal to the product of edge lengths, we can thus geometrically define the volume of a slice $S_i$ defined by subsets $\{s_{i,j}\}$ as $V(S_i)=\prod_{j=1}^p L(s_{i,j})$; this is equivalent to only considering the product of lengths for features that define it, since the others have length 1.  Slice volumes thus also satisfy $0<V(S_i)\leq 1$.  For the full feature space, $V(\mathcal{S})=\prod_{j=1}^p L(\textrm{dom}(F_j))=\prod_{j=1}^p 1 =1$.  If a set of slices $\{S_i\}$ fully partition $\mathcal{S}$, then $\sum_i V(S_i)=V(\mathcal{S})=1$.  Appendix~\ref{ssec:volume_example} shows a full example of a density partition, specification of slices $\{S_i\}$, calculation of lengths of their defining subsets $\{s_{i,j}\}$ and volumes $\{V(S_i)\}$, and that the volumes sum to 1.

Realistic dataset features typically have non-uniform marginal distributions and
inter-feature dependence (e.g., correlations and domain feature constraints).  For instance, we'd expect the feature $F_j=\textrm{STATE}$ to have a non-uniform distribution over U.S. states, since states have very variable population sizes.  A feature $F_j=\textrm{INCOME}$ would also likely have a non-uniform distribution, since income distributions are often skewed, with very high incomes (which stretch the domain boundaries) being very rare.  For features HEIGHT and WEIGHT on a dataset of humans, it is rare to see very heavy short people or very light tall people (where ``very" here means in the upper range of the marginal observed values of a feature, disregarding the other features); these are feature constraints that are soft, but certain feature value combinations may be impossible in the domain (e.g., an attorney without a law degree).  These give rise to regions (which we model by hyper-rectangles) that have sparse density or are empty.  Realistic datasets thus should have varying feature density throughout the potential space $\mathcal{S}$, which makes it logical to partition $\mathcal{S}$ by density.  See Section~\ref{sec:metrics} for an example of a summary metric that can address identifying when $D$ seems to have too much uniformity and mutual independence, which would be unrealistic.  
\section{Algorithm Overview
\label{sec:algorithm}}

Here, we outline the elements of our method, which consists of the following steps, applied iteratively:

\begin{enumerate}
    \item Calculation of a numeric target $\mathbf{y}=\{y_i\}_{i=1}^n$ that serves as a proxy for the density of observation $i$.
    \item Use regression trees with target $\mathbf{y}$ to partition the feature space $\mathcal{S}$ on features $F_1,\dots,F_p$.  
    \item Carve out empty space along the way from the newest resulting split.
\end{enumerate}

\subsection{Density proxy determination
\label{ssec:density_proxy}}

Our method requires a numeric target $y_i$ for each observation $\mathbf{x}_i$, which represents this observation's multivariate density within the feature space $\mathcal{S}$, similarly to kernel density estimation (KDE) or the methods in density-based clustering, such as DBSCAN.  Observations with high density should have many other points within a small neighborhood, while sparse points should have relatively few neighborhoods or be surrounded by more empty space, similarly to how KDE estimates a point's density by distances to all or some of its neighbors.  As discussed in Section~\ref{ssec:region_partition}, $y_i$ will then serve as the target for partition.  As such, it should be approximately monotonically increasing or decreasing with the density of $\mathbf{x}_i$; that is, a higher value can represent higher density or more anomalousness, which should correspond to lower density.

There are several methods that can be used.  One is to use a distance-based clustering algorithm, such as OPTICS (Ordering Points To Identify the Clustering Structure, \cite{ABKS1999}), where a numeric output such as the core distance of $\mathbf{x}_i$ can be used as $y_i$; the cluster identifications are not used.  In OPTICS, the core distance is the distance from $\mathbf{x}_i$ to its $\nth{m}$ closest neighbor, where $m$ is the (user-specified) minimum number of observations within an $\epsilon$-radius neighborhood of $\mathbf{x}_i$ for it to be considered a core point.  Thus, a lower core distance should indicate a higher density of $\mathbf{x}_i$.   Gower distance (\cite{G1971}) is a metric which calculates the multivariate distance between observations $\mathbf{x}_i$ and $\mathbf{x}_j$ as the average of their feature-wise distances.  The distances can be tailored to the feature type (e.g., range-normalized Manhattan distance for numeric, Dice coefficient for nominal categorical, or Manhattan distance for ordered categorical; see \cite{A2020}), which means it can apply to mixed data types.  If Gower distance is used as the distance metric in, say, OPTICS, the core distance (i.e., $y_i$) can be considered as a distance-based density measure of observation $\mathbf{x}_i$.

Depending on the implementation, clustering-based methods can also scale poorly with $n$ in terms of computational complexity.  An alternative to obtaining a proxy $y_i$ by clustering with an appropriate distance-based metric is to use an anomaly score; here, a higher anomaly score should correspond to lower density, but it may not correspond directly or proportionately to distance-based sparsity.  One such anomaly scoring method is isolation forests (IF, \cite{LTZ2008}), which build a forest (ensemble) of trees on subsets of the features; the trees perform binary splits on the ranges of the features to isolate observations.  The more splits required to isolate an observation $\mathbf{x}_i$, the more anomalous it is; the anomaly score (normalized to $[0,1]$) can be used as $y_i$, so a higher score indicates lower density.  IF is very fast and computationally light. An extension of IF (\cite{C2019a}, implemented for Python as \cite{C2019b}), corrects some of the weaknesses of \cite{LTZ2008} in splits on categorical variables, which tend to have lower weight than numeric variables in determining the anomaly scores, due to their typical lower number of unique values.

It is important to note that since the feature space $\mathcal{S}=\bigcap_{j=1}^p\textrm{dom}(F_j)$ is the bounding hyper-rectangle of all observations in $D$, its definition is sensitive to outliers if they affect the boundary points of the domain of a feature.  For instance, say $F_1$ is INCOME, and the current domain is $[\$0,\:\$200{,}000]$; if a new observation is added with $F_1=\$1{,}000{,}000$, the $\textrm{dom}(F_1)$ now grows five-fold.  Assuming none of the other feature domains are affected, $\mathcal{S}$ now grows five-fold along the $F_1$ dimension.  Since $V(\mathcal{S})$ must always be 1, this means this single observation has created an empty region $S=\{\$200{,}000 <\:\textrm{INCOME}\:<\$1{,}000{,}000\}$ of volume approximately 0.8, and so the non-empty slices built on $D$ previously would now shrink by approximately a factor of 5.

Such outliers will tend to receive a score $y_i$ that indicates high sparsity or anomalousness.  To make the partition more robust, it may be wise to omit the observations with, say, the highest 1\% of sparsity scores before defining $\mathcal{S}$ and conducting the partition.  If this million-dollar income observation is unique in the dataset, including it in the partition may make the results non-robust, and so it may be dropped.  However, say that 5\% of the observations have income of $\$1{,}000{,}000$, and the next highest income is $\$200{,}000$, these high earners will likely be neighbors of each other in $\mathcal{S}$, giving them less extreme density targets than otherwise.  Even though they are unusual relative to the other observations, some of them will likely be included in $D$ even if, say, the sparsest 1\% are trimmed.  Trimming the sparsest observations can affect observations not on the boundaries of $\mathcal{S}$ if, say, they are surrounded by relatively empty space.  In this case, trimming them can give a more parsimonious representation of the empty space than if the partition has to `cut around' these observations.  This is a similar issue to the general decision of how many outliers to trim from a sample when estimating the population distribution to make the estimate robust to outliers; if points that are unusual but not very extreme are trimmed, the resulting estimate may not be accurate since these unusual points in fact do describe the distribution.

\subsection{Region partitioning
\label{ssec:region_partition}}

Once we have a have a numeric target $\mathbf{y}$ which serves as a proxy for the density, a density-based partition of $\mathcal{S}$ can be obtained by a model mapping $D\rightarrow \mathbf{y}$.  If the model performs hierarchical binary splits on the input features $F_1,\dots,F_p$, this yields interpretable rectangular slices (as defined in Section~\ref{sec:problem}) on these features.

One such widely-used techniques are regression trees (\cite{BFSO1984}).  Regression trees construct binary trees on the input features, which at each node conduct a binary split on one of the features (or a one-hot encoding column corresponding to one level of a nominal categorical feature) such that the mean squared error (or a similar metric) of the numeric target $\{\mathbf{y}_i\}$ for observations $\mathbf{x}_i$ is minimized given the choice of split on the range of the chosen feature $F_j$.

\begin{figure}
    \centering
    \includegraphics[scale=0.6]{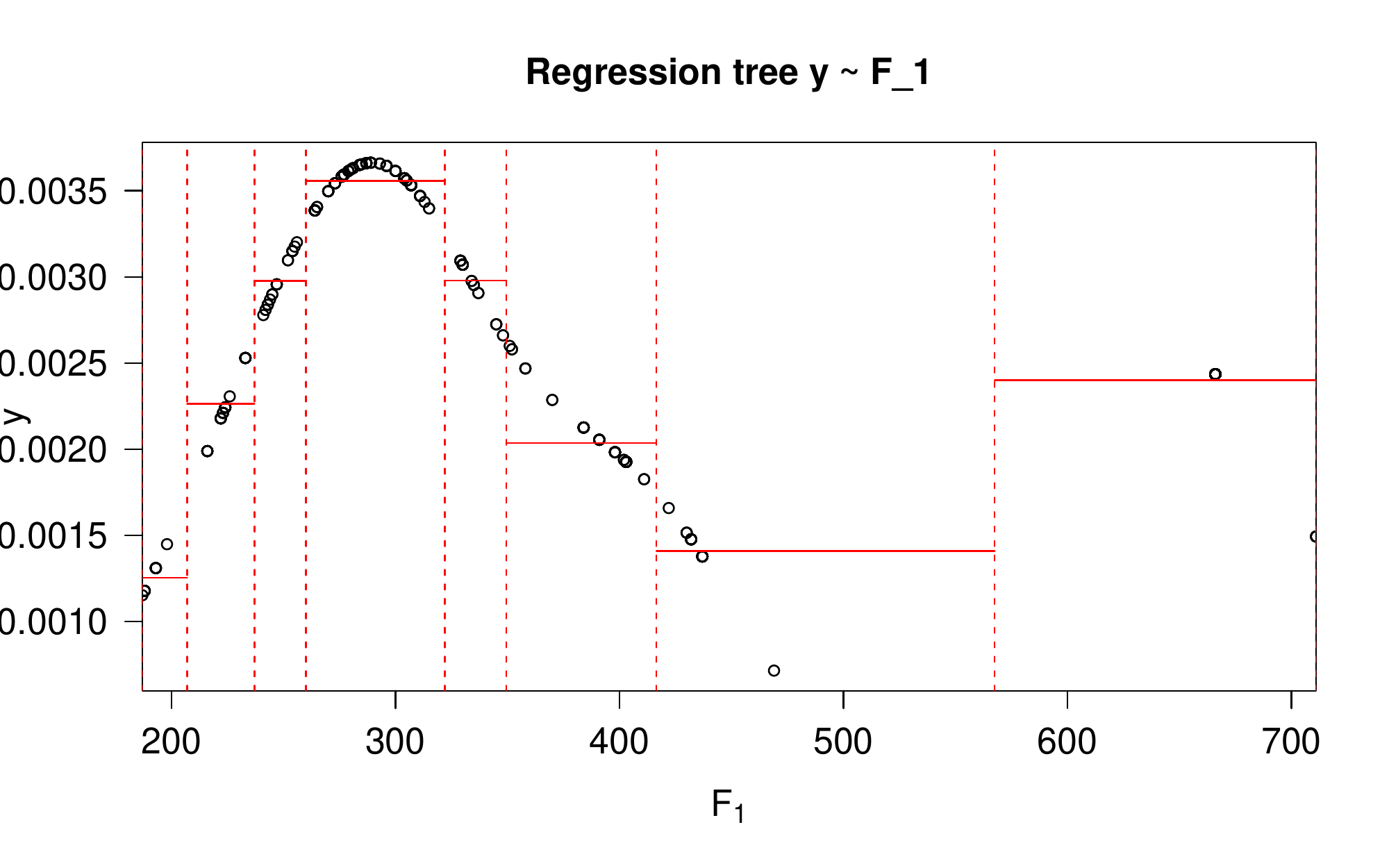}
    \caption{\label{fig:regression_tree} Regression tree on a univariate numeric feature $F_1$ (horizontal axis) and numeric target $\mathbf{y}$, in this case the kernel density estimate.
}
\end{figure}

An illustration is shown in Figure~\ref{fig:regression_tree}, where a regression tree maps a univariate feature $\mathbf{x}=F_1$ to a numeric target $\mathbf{y}$, where $y_i=f(x_i)$ is the KDE of $F_1$ at a given value $x_i$.  Each vertical dashed line is a split on $F_1$ in the tree, and the interval between each pair of consecutive dashed vertical lines constitutes a region, or slice (see Section~\ref{sec:problem}).  For instance, here $\textrm{dom}(F_1)=[187,\:711]$ and we have 8 slices, where $S_1=\{187\:\leq F_1\leq \: 207\}$ and $S_8=\{567.5\:< F_1\leq \: 711\}$.  Within each slice, the solid horizontal line is $\textrm{E}(y_i\mid \mathbf{x}_i\in S_j)$, the average value of $y_i$ in each slice.  Since $\mathbf{y}$ is the KDE, this average reflects the average density of the observations, and thus follows the shape of the KDE curve.  In this way, the set of slices $\{S_j\}_{j=1}^8$ represents a density-based partition of the observed span $\textrm{dom}(F_1)$.

\subsection{Carving out empty regions
\label{ssec:empty_space}}

One major aspect of our method is its ability to carve out slices that represent empty space in $\mathcal{S}$.  We currently search for empty space in two locations:  
\begin{enumerate}
    \item Only on the feature and split value used by the regression tree at that step (Section~\ref{sssec:empty_space_at_split}).
    \item After splitting, at any feature and location of the `outside' of the observed points in that slice (Section~\ref{sssec:empty_space_after_split}).
\end{enumerate}

Another heuristic would be to find internal empty space in a given slice, which would split the slice into at least one empty slice, at least two new non-empty slices.  However, we do not currently implement this.

\subsubsection{Empty regions at regression tree splits \label{sssec:empty_space_at_split}}
\begin{figure}
    \centering
    \includegraphics[scale=0.6]{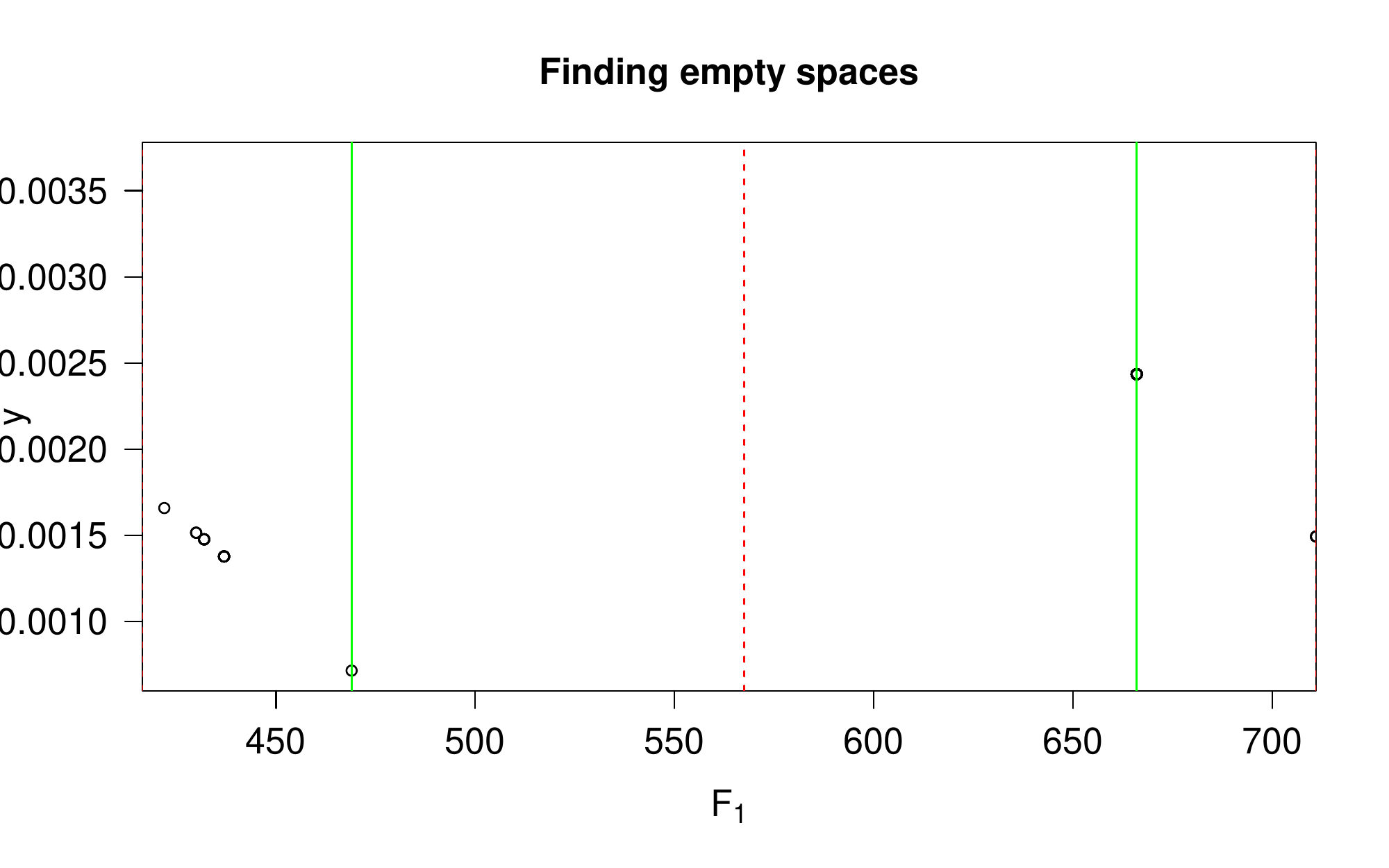}
    \caption{\label{fig:empty_space}
    Empty space (between green lines) around split (red dashed line) between rightmost slices $S_7$ and $S_8$ in Figure~\ref{fig:regression_tree}.
    }
\end{figure}

An illustrative example is shown in Figure~\ref{fig:empty_space}.
  In Figure~\ref{fig:regression_tree}, the rightmost two slices are $S_7=\{416.5\:<F_1 \leq \: 567.5\}$ and $S_8=\{567.5\:< F_1\leq \: 711\}$.   However, the interval $(469,\: 666)$, indicated by the green vertical lines in Figure~\ref{fig:empty_space}, is empty, which is useful information to the user.  The split points (e.g., $F_1=567.5$) are the mid-points between the highest and lowest values in the slices on the left and right, respectively, and the splitting decision is typically greedy in that it does not account for empty space in the observed values of numeric features $F_j$.  The length (see Appendix~\ref{ssec:edge_length}) of this empty interval is $0.3755782$ of the overall domain.  

Our method uses a heuristic that empty space, such as $(469,\:666)$ above, which straddles the location of a regression tree split, will only be carved out if its length $L(\cdot) >\textrm{min}_L$, for a user-specified value $0\leq \textrm{min}_L\leq 1$.  Setting $\textrm{min}_L=1$---that is, the maximum possible---means no empty space will ever be carved out, and so the partition will remain as in Figure~\ref{fig:regression_tree}, where each slice contains observations.  Setting $\textrm{min}_L=0$ means all possible empty space (subject to the limit $p^*$ on slice dimension), no matter how small, will be carved out, which over-fits the partition to the observed data, in which case the empty regions may not represent real feature constraints, but rather artefacts of the data.  We recommend setting $\textrm{min}_L=0.1$, for instance (see example Figure~\ref{fig:scatterplots}).  In that case, the empty space in Figure~\ref{fig:empty_space} would be large enough, and so the resulting partition would have $S_1,\dots,S_6$ as in Figure~\ref{fig:regression_tree}, but with $S_7=\{416.5\:<F_1 \leq \: 469\}$, an empty slice $S_8=\{469\:<F_1 < \: 666\}$  and $S_9=\{666\:\leq F_1\leq \: 711\}$.  The example shown here is on numeric data, but the same procedure applies to nominal categorical features, as discussed below.

\subsubsection{Empty regions after regression splits \label{sssec:empty_space_after_split}}

After carving out empty space from splits, we also employ a heuristic to carve out empty space anywhere on the `outside' of a given slice $S_k$ after it is formed from a regression tree split, as opposed to specifically at the regression tree split feature and value.  The method determines the empty space between the boundaries of $S_k$ and the boundaries of the observed $\{\mathbf{x}_i\colon\:\mathbf{x}_i\in S_k\}$.  For instance, in the above, $S_2=\{207\:<F_1 \leq \: 237\}$, but the observed values span only $[216, 233]$; the empty boundary space for this feature in slice $S_2$ is $s'_{2,1}=(207,216)\bigcup(233,237]$; for real-valued, integer-valued, or ordered categorical features, this empty space can be a union of two sets.  If $p\geq 2$, this trimming can occur on any of the features, while the first form of trimming occurs only on the feature used to split the nodes, and only at that split threshold.

The (potentially zero-size) boundary gaps $s'_{k,j}$ in slice $S_k$ are calculated for each feature $F_j,\: j=1,\dots,p$.  If any have $L(s'_{k,j})>\textrm{min}_L$, these are iteratively trimmed in decreasing order of size, resulting in an empty space of $S_k \bigcap s'_{k,j}$, and $S_k$ is re-defined as $S_k\setminus (S_k \bigcap s'_{k,j})$, subtracting the empty space.  That is, the interval $s_{k,j}$ which defines $S_k$ on feature $F_j$ is redefined as $s_{k,j}\setminus s'_{k,j}$  For instance, $S_2$ could then be re-defined as
the smaller $S_2=\{216 \leq\: F_1\:\leq 233\}$, and two new empty slices, $\{207 < \:F_1\: < 216\}$
and $\{233 <\: F1 \leq 237\}$ would be added, since the empty gap $S_2\bigcap s'_{2,1}$ consists of non-contiguous
intervals. 

In each iteration, before trimming, the slice $S_k=\bigcap_{j=1}^p s_{k,j}$.  Carving on $F_j$ results in a trimmed slice $S_k=(\bigcap_{i\ne j}s_{k,i})\bigcap(s_{k,j}\setminus s'_{k,j})$ and one or two (if $s'_{k,j}$ is non-continguous) empty slice(s) $(\bigcap_{i\ne j}s_{k,i})\bigcap(s'_{k,j})$, which have the same definition on all features except $j$.  If, before trimming $s_{k,j}\ne \textrm{dom}(F_j)$, then $S_k$ was already defined on $F_j$.  Hence, slice $S_k$'s dimension will not increase after trimming, and will also equal the dimension of the empty slice; otherwise, if $s_{k,j}=\textrm{dom}(F_j)$ before trimming, its dimension would increase by 1, and so would the empty space.  Thus, for trimming to occur at each iteration, two conditions must be met:
\begin{enumerate}
    \item The dimension of $S_k$ must remain $\leq p^*$ after trimming.  That is, it must have either been $<p^*$, or, if the dimension was $=p^*$, it must have been defined on $F_j$. See Section~\ref{ssec:feature_domains}.
    \item The resulting empty slice must be large enough on all dimensions.  That is, we must have $L(s_{k,i})>\textrm{min}_L,\:\forall i\ne j$, and $L(s'_{k,j})>\textrm{min}_L$ as well.  This restriction applies only to the empty space, and recall that no such restriction is put on the regression tree when forming the slices initially.  If, for instance, the resulting $S_k$ is re-defined such that $L(s_{k,j}\setminus s'_{k,j})\leq \textrm{min}_L$ (it is `narrow' along feature $F_j$), this is fine.
\end{enumerate}

In the example of $S_2$ above defined only on $F_1$, $L(s'_{2,1})$ is too small to trim, so $S_2$ is left as is.

Note that the for simplicity, the illustrations of regression trees and empty space carving have used a univariate numeric case, but the same calculations of empty space carving can be done on categorical features as well.  For instance, say there are two features $F_1$ and $F_2$, where $F_2$ is RACE, with $\textrm{dom}(F_2)=\{\textrm{White},\textrm{Black},\textrm{Asian}\}$.  Say, $\mathcal{S}$ is first partitioned only on $F_1$ into slices $S_1$ and $S_2$; that is, $S_1=s_{1,1}$ and $S_2=s_{2,1}$, which are complementary subsets of $\textrm{dom}(F_1)$.  Even though $S_1$ is not defined on $F_2$ (i.e., currently $s_{1,2}=\textrm{dom}(F_2)$), say $S_1$ only contains observations with $F_2\in\{\textrm{Black},\textrm{White}\}$ (all Asian individuals are in $S_2$).  There is therefore an empty gap in this slice, $s'_{1,2}=\{F_2\in\{\textrm{Asian}\}\}$, where the complement is $s_{1,2}\setminus s'_{1,2}=\{\textrm{White},\textrm{Black}\}$; that is, $S_1$ could be defined on $F_2$ as well, since the data it contains do not span the full $\textrm{dom}(F_2)$.  

This gap has length $L(s'_{1,2})=1/3$, since it contains one of three possible values of RACE, and forms an 2-dimensional empty slice defined as $S_3=S_1\bigcap s'_{1,2}$.  If the maximum dimension $p^*>1$, and if $S_3$ is large enough on all sides (both $L(s_{1,2}),L(s'_{1,1}) > \textrm{min}_L$), then $S_3$ (Asian individuals satisfying $S_1$, which are unobserved in $D$) becomes a new empty slice, defined as $S_3=s_{3,1}\bigcap s_{3,2}$, where $s_{3,1}=s_{1,1}$ and $s_{3,2}=s'_{1,2}$. $S_1$ is re-defined as $S_1\setminus(S_1\bigcap s'_{1,2})=s_{1,1}\bigcap (s_{1,2}\setminus s'_{1,2})$ (narrowing $S_1$ to omit Asian individuals, a combination that is not observed in $D$).

\section{Alternative methods
\label{sec:alternatives}}

As mentioned in Section~\ref{sec:problem}, the main aspects of our method are
\begin{enumerate}
    \item Allows density to be characterized on features of mixed type, without discretization, gridding, or encoding that distorts the original feature values.
    \item Returns regions in the form of interpretable hyper-rectangles.
    \item Returns regions that are empty (contain no observed features) in addition to those with data.
\end{enumerate}

There are other methods that characterize an observed data space by density.  ML clustering methods, by definition, typically only characterize groups of observed data points, and not empty space (item 3).  This includes methods like DBSCAN (\cite{EKSX1996}) and others.  DBSCAN and related methods (HDBSCAN, etc.) can identify density-based cluster patterns that can be of arbitrary shapes (e.g., not spherical or rectangular) that capture visual patterns that can be discerned in 2-D space; by extension, they should also work in higher-dimension space and with non-Euclidean distance metrics.  While they may perform clustering well on difficult problems, in our application we actually want clusters of pre-defined shape types (rectangular) and not of arbitrary type (item 2).

There are several similar methods to ours that do match on item 2, and return hyper-rectangular regions.  One is \cite{OONE1999}; however, this requires numeric features only, as well as pre-gridding (item 1).  Another method (\cite{CM2003}, a patent filed by Oracle International Group), can work on mixed feature types, and also includes methods to build and update partitions with new data.  It does pre-grid numeric data, but we note that this method also addresses several concerns that ours does not, particularly the ability to update a partition on a dataset when new data arrive without having to re-run the partitioning on the combined data.

The method in \cite{CM2003} differs in two major ways from ours in terms of the split decision.  First, their decision to split a region on a feature axis is made on a marginal (individual feature axis) basis based on that feature's distribution as the target.  As discussed, our method uses a density target (Section~\ref{ssec:density_proxy}) that uses all features' values which may obscure differences in the marginal distributions, particularly when the feature dimension is high.  Secondly, their criterion for selecting a split point in a histogram so it results in two regions that differ maximally by density.  The criterion is  ``selecting a splitting plane at a valley, which is a bin of low density in the histogram, between two peaks, which are bins of high density and wherein a difference between peak and valley histogram
counts are statistically significant" (\cite{CM2003}, page 20).  This method splits on a valley surrounded by two peaks, an ideal bi-modal scenario.  Assuming this is to be interpreted strictly, we note that our method does not make this assumption.  Instead, since the splitting decisions are based on the regression tree, our method can split on cases where there is a uni-modal density, such as a peak with flat areas (`valleys' without a corresponding peak on the other side) on one or both sides; the first case would resemble a monotonic increasing or decreasing density scenario, such as a step function.
\section{Examples
\label{sec:examples}}

For the following, we use the Adult dataset (\cite{KB1996}), a subset of records of respondents to the U.S. Census conducted in 1994; the illustrations use a $n=1,000$-observation subset and only the $F_1=\textrm{AGE}$ and $F_2=\textrm{HOURS\_PER\_WEEK}$ (hours worked per week) features.  Note, the dataset was filtered to omit non-working respondents, that is, with $\textrm{HOURS\_PER\_WEEK}=0$.  Thus, $p=2$ and we set $p^*=2$ as well to allow partitions to be made on both features; both features are also coded as integer-valued (see Section~\ref{ssec:feature_domains}).  An additional parameter \texttt{min\_slice\_size\_frac}, not discussed earlier, is set to 0.1, meaning the regression tree will not form a leaf (i.e., a potential partition slice) if contains less than 10\% of the dataset observations ($0.1n$), with a minimum size of 2 observations; this is a form of robustness control on the partition.

\begin{figure}
    \centering
    \includegraphics[scale=0.6]{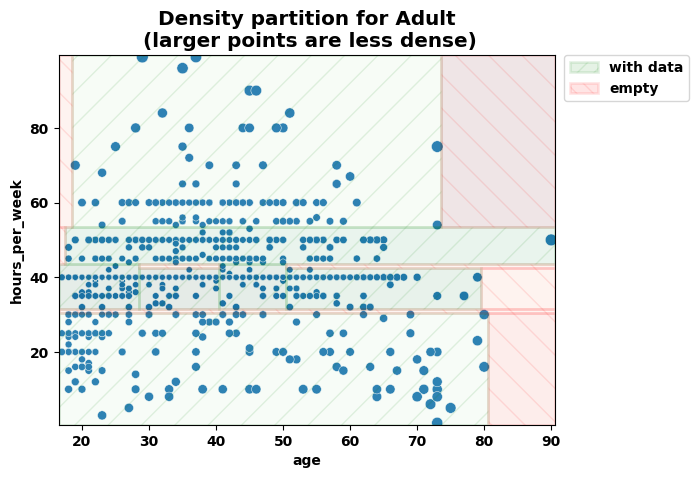}
    \includegraphics[scale=0.6]{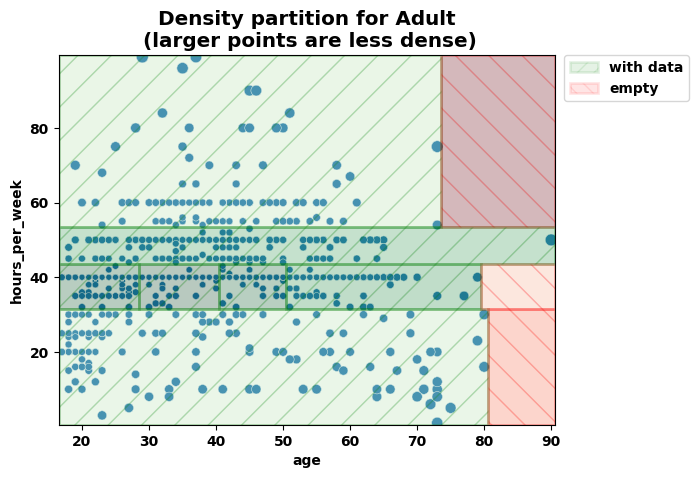}           \includegraphics[scale=0.6]{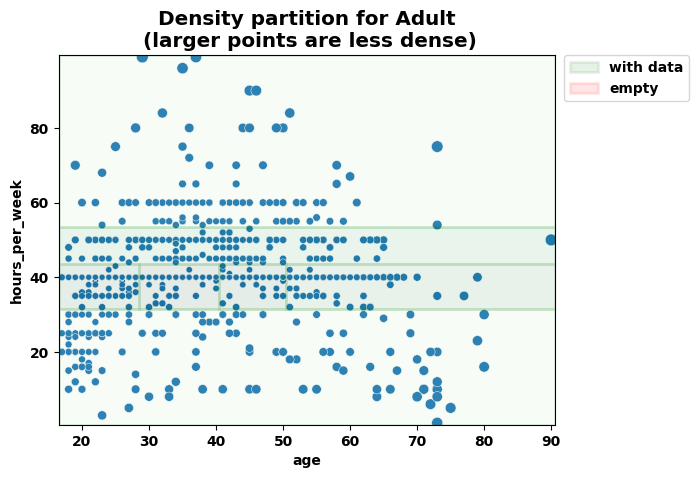}
    \caption{\label{fig:scatterplots}
    Density-based partition of $n=1000$ observations from `Adult' dataset on $F_1=\textrm{AGE}$ and $F_2=\textrm{HOURS\_PER\_WEEK}$.  The plots show, from top to bottom, results with $\textrm{min}_L=0.0,\:0.1,\:\textrm{and}\:1.0$.
}
\end{figure}

Figure~\ref{fig:scatterplots} show the visualization of a scatterplot of the dataset observations, with an overlay of rectangles representing the resulting partition. Seven non-empty slices are found for each, but the partitions differ in the amount of empty space carved out from these slices to form new empty slices, so the total number of slices can differ. Empty space slices are shaded in red, with the hue darkening with increasing volume; those with data are shaded in green, with the hue darkening with increasing average density (decreasing anomalousness or core distance).  Shadings are only comparable for partitions within a plot, not between plots.  The plots are shown for values of $\textrm{min}_L=0.0,\: 0.1,\:\textrm{and}\:1.0$ (see Section~\ref{ssec:region_partition}), which controls the amount of empty space carved out.  As noted, the top plot, with $\textrm{min}_L=0.0$, results in many small empty (red) partitions, since all empty spaces of any size are carved out; the resulting partition is unlikely to be very robust to, say, another sample similar to $D$, due to the overfitting.  In the bottom plot, on the other extreme, no empty spaces are carved out.  One-dimensional slices are rectangles for which one side occupies the entire axis (such as the rightmost slice in this plot), and two-dimensional slices have both sides being shorter than the respective axes.

In the middle plot, with $\textrm{min}_L=0.1$, only three empty slices, in the upper and lower right hand corners, are found. These represent a constraint that, in the given dataset, at least, that workers are unlikely to be older (recall non-workers are omitted).  Furthermore, the upper right hand empty space is wider along the AGE axis, indicating that older workers (about, say their mid-70s) are unlikely to work a high number of hours (above, say, 55) per week.  Note that there is a single unusual respondent aged 90, who is significantly older than the next oldest respondent, aged 80, and retaining such outliers can alter the configuration of slices.  In addition to this outlier, there are several outliers in the upper portion of the plot, indicating respondents working a very high number of hours (more than 90).  The more dense (smaller points), more `typical' observations are in the center of $\mathcal{S}$, containing respondents of typical working ages (20s-60s) working typical workweeks ($\approx30-60$ hours per week).  Here, the slices tend to have smaller volume to fit the bulk of the data better.  The plot would look cleaner if two empty slices in the bottom right corner were merged, but we do not conduct any such post-processing, and the empty spaces are a byproduct of the partition on observed data.

\section{Analysis metrics
\label{sec:metrics}}

Once a density partition $S_1,\dots,S_k$ is made of $\mathcal{S}$, we can perform some calculations to summarize the results.  As noted in Section~\ref{sec:problem}, a realistic dataset should have unevenly-distributed points within $\mathcal{S}$, and these partitions by density, in addition to empty spaces, if found, should characterize the domain constraints of the features.  Hence, a realistic dataset---or a synthetic dataset generated to have realistic, and not independent, inter-feature associations---should have slices of various volumes and densities.  In addition to summarizing the distribution of observation density in a single dataset, we may also want to compare the distributions of two different datasets.  

Say a partition on $D$ results in $K$ slices.
Let $\phi(S_k)=\frac{\sum_{i=1}^n I(\mathbf{x}_i\in S_k)}{n},\:k=1,\dots,K$ be the fraction of the observations in $D$ contained in slice $S_k$; $\phi_k=0$ if $S_k$ is empty space.
If the distribution of observations in $D$ is perfectly uniform within the feature space $\mathcal{S}$, each slice $S_k$ should have $V(S_k)=\phi(S_k)$.  Slices $S_k$ that are denser than average should have $\phi(S_k)>V(S_k)$; that is, the region covered by $S_k$ contains a higher fraction of observations than its volume (which is a fraction of the total feature space volume).  Recall that both $\sum_k \phi(S_k)=\sum_k V(S_k)=1$.  This indicates that a chi-squared statistic

\[\chi(D, \{S_k\})=\sum_{k=1}^K\frac{(\phi(S_k) - V(S_k))^2}{V(S_k)}
\]

where the statistic $\chi(D, \{S_k\})$ follows a chi-squared distribution with $K-1$ degrees of freedom ($\chi^2_{K-1}$).  A perfectly uniform dataset $D$ should have $\chi(D, \{S_k\})\approx 0$; in the most extreme case, the algorithm will not be able to generate a partition (detecting no variances in density), and thus $K=1$ with $S_1=\mathcal{S}$, and $V(S_1)=\phi(S_1)=1$.  The volumes are set as the expected values of $\phi(S_k)$ under the null hypothesis of uniformity since $V(S_k)>0,\:\forall k$, while $\phi(S_k)$ may equal 0.  The p-value of this statistic, according to the chi-squared distribution, can measure the likelihood the distribution is non-uniform, and the value $\frac{\chi(D, \{S_k\})}{K-1}$ can be used as a simple metric to compare uniformity of density between different datasets $D$ and $D'$ by their respective density partitions, where higher values indicate less uniformity.

In addition, we recommend requiring the minimum decrease in MSE by splitting a node to be above a minimum threshold, such as a fixed percentage of the overall variance in $\mathbf{y}$.  This will help control the robustness of the partition to small changes in density, due to the greediness of the regression tree splitting.  As noted in Section~\ref{ssec:density_proxy}, extreme outliers should be trimmed before such an analysis, otherwise $\chi$ would be too sensitive to outliers, due to changes in the volume of empty space in $\mathcal{S}$.



\section{Further applications
\label{sec:applications}}

Beyond simply constructing a density-based partition on a dataset, our method has potential applications to other data analysis tasks, as discussed below.

\subsection{Identification of areas of low classifier performance
\label{ssec:error_identification}}

A related work to that presented here is a tool called Frea-AI (\cite{ARZ2020}).  This tool similarly uses decision trees to identify low-dimensional slices of observations where a classifier model has lower predictive accuracy than it does on $D$ overall; these areas are `weak' in terms of the classifier's performance.  The slices in Frea-AI are of the same form as the ones here.  Since the decision tree suffers from the same greediness feature as regression trees (see Section~\ref{ssec:region_partition}) in that the compactness (i.e., observation density) of the tree nodes is not considered, we have noted that sometimes the Frea error-based slices span too wide a range, and include some points that are in sparse regions of $\mathcal{S}$.  We are currently investigating whether mis-classified observations, as compared to correctly-classified ones, tend to fall in sparser or denser areas.  

However, it is possible that the error-based slicing results could be improved or supplemented by a density-based analysis, such as including density in the optimization criterion, or by restricting the search for error-based slices to the denser regions of $\mathcal{S}$.  This is because predictions on observations $\mathbf{x}$ made in very sparse regions of $\mathcal{S}$ may be inherently unreliable, and thus such observations which have classification errors may not be due to deficiencies in the classifer. The user may be more concerned about identifying error-prone regions in dense areas of $\mathcal{S}$.  For instance, if an ML model predicts a consumer's spending based on the individual's characteristics, it may be more worrisome if it makes errors on middle-class consumers (a more dense area) than on millionaires (likely in sparser areas of $\mathcal{S}$).

Another potential use in modeling classifier error is to infer from partitions $\{S_i\}$ on a dataset $D$ to a similar dataset $D'$ which may be unlabeled, and thus to infer about the classification error on it.  First, it may be reasonable to assume that the classification error for observations in $D'$ in a given slice $S_i$ may be similar to that observed in $S_i$ for observations in $D$.  Second, if a given empty slice $S_i$ in $D$ is non-empty in $D'$, perhaps the classification error rate for these observations is likely to be relatively high.  These two facts may help either approximate or set a conservative bound on the expected classification error rate on $D'$.

\subsection{Causal inference
\label{ssec:causal_inference}}
In the example below, we see whether our designation of slices that are empty or of sparse density can indicate groups of observations (feature value combinations) for which a causal analysis of comparison of outcome effects between treatment levels cannot be effectively conducted.

\subsubsection{Background}

Causal inference is a technique used, particularly in observational studies, to infer the effect of a categorical ``treatment" variable on some (typically numeric) outcome of interest.  The effect is measured by comparing the difference in average outcome value between groups with different treatments. For instance, one may want to assess the difference in life expectancy (outcome) between smokers and non-smokers (``smoking" is a binary treatment variable).  In the dataset we will use, the authors want to estimate the effect of participating in a work training program (TRAINING=1 means they received the treatment of training) on income (in 1978); hence, the average 1978 incomes of individuals receiving or not receiving training will be compared.  
 
In randomized clinical trials (RCTs) to, say, estimate the effect of a given drug treatment on cholesterol levels (outcome), participants may be randomly allocated to receive either the drug or a placebo (control).  Each of these groups, whether they received a drug, placebo, or some other control, is a treatment group.  The randomization is intended so that for any feature of interest (e.g., age, race, sex, health conditions), when comparing the average outcome between the treatment groups (typically the drug at different levels and the control vs each other), the effect should only be due to the difference in the treatment.  However, certain instances, treatment randomization cannot be done.  For instance, it is unethical to randomly assign people to smoke, or perhaps we wish to study some historical phenomenon; so, we often have to rely on observational data or studies.  This means that comparing the average outcome difference may reflect other confounding factors; for instance, people who smoke may have lower life expectancy because they also tend to have certain other health conditions or behaviors, which non-smokers display to a different degree, which also affect the outcome.  
 
In a causal inference analysis, we require that various feature combinations (e.g. males between 30 $<$ age $<$ 50) display ``positivity", that is, that they have been observed enough in both the treatment and non-treatment groups, so we can compare their averages. One way to deal with this in causal inference is to make the treatment and non-treatment groups more similar along certain feature dimensions. For instance, the propensity score, which reflects the likelihood of an individual being in the (not randomly assigned) treatment group, given various dataset features, is calculated. Individuals with certain feature combinations (e.g., certain ``slices") with very low average propensity, are unlikely to be in the treatment group; such groups are said to ``violate" the positivity assumption. A causal inference analysis will consist of comparing individuals' outcome under the treatment they received (in this case, training or not) with the estimated outcome under treatments they didn't receive.  But if their average propensity score is low, this means we may have very little (or zero) data (observations) of such people in one of the treatment groups, and hence these individuals (in the form of slices grouping these individuals along common feature values) should be removed from the data before re-calculating the average difference in outcome.\footnote{Readers interested in more details about causal inference are referred to \cite{N2020}; video slides are available at \url{https://www.bradyneal.com/causal-inference-course}}

In \cite{SKD2020}, an example of a causal inference analysis using Python's \texttt{causallib} module (\cite{SKDNKW2019}, see also \cite{SKRBNAMG2019}), on the Lalonde dataset, is shown. 
The Lalonde dataset is used for estimating the effect of participating in a work training program on earnings of individuals in 1978. It consists of the original participants in the experiment, who were randomly assigned to either receive training or not, supplemented by a random draw of individuals from a population-level survey, to serve as additional control (non-training) observations. The numeric features used in the dataset are \{AGE, EDUCATION, RE74, RE75, RE78\}, where RE74, RE75, and RE78 are the earnings in 1974, 1975 and 1978 correspondingly. The categorical features used are \{TRAINING, BLACK, HISPANIC, MARRIED, NO-DEGREE\}, all of which are binary indicators. In this case, the treatment indicator is TRAINING (binary positive if received training), and the outcome variable is RE78 (numeric earnings received in 1978).  This dataset is very imbalanced on the treatment variable, it has 22,106 records in total, out of which only 185 received training.

Direct causal inference analysis for this dataset, without adjustment, isn't successful due to high percentage of positivity-violating records. To overcome this, the propensity score is estimated, and a binary ``violation" indicator is set to 1 for all records with propensity score below a certain threshold. Then a decision tree based algorithm is proposed to find ``slices" (i.e. hyper-rectangles) with high percentage of violating records (for  example, the slice defined by $\{\textrm{AGE}\:<55\} \:\&\: \{\textrm{BLACK} \in\{\textrm{False}\}\} \:\&\: \{\textrm{RE75}\:>10{,}000\}$ is found), and once removed, the causal inference yields much better results.

\begin{table}[ht]
\centering
\begin{tabular}{ |c c c|c|c|c|} 
\hline
\multicolumn{3}{|c|}{Feature Rules} & \multicolumn{2}{c|}{Treated} & \multicolumn{1}{c|}{Size}\\
\cline{4-5}

 \multicolumn{3}{|c|}{} &True & False &\\
  \hline
  $\textrm{BLACK}\in\{\textrm{False}\}$   & RE75 $>$ 10K & AGE $>$ 21 &0 &11056 &11056\\ 
  \hline
 $\textrm{BLACK}\in\{\textrm{False}\}$   & RE75 $>$ 5.5K  & $\textrm{MARRIED}\in\{\textrm{True}\}$ &1 &11049 & 11050 \\ 
  \hline
  $\textrm{BLACK}\in\{\textrm{False}\}$   & RE75 $>$ 13K  & AGE $>$ 40 &0 &6373 &6373 \\ 
  \hline
  $\textrm{BLACK}\in\{\textrm{False}\}$   & RE74 $>$ 0K  & AGE $>$ 39  &0 &5096 &5096\\ 
  \hline
  RE74 $>$ 20K   & RE75 $>$ 15K  & AGE $>$ 40 &0&2707 &2707 \\ 
  \hline
\end{tabular}
\caption{\label{tab:DT_violating_slices} List of slices removed in \cite{SKDNKW2019}.}
\end{table}
Table~\ref{tab:DT_violating_slices} describes the violating slices found and removed in the work.
The effect calculated before finding and removing the slices was $-\$2,658$. The negativity of this initial result is contradictory, as it implies that participants in the training program are estimated to earn less than those who did not receive training. After removing the violating slices the effect rose to $\$567$; the fact this is positive rather than negative suggests it is a more logical result.

Here we propose a method for finding slices with high percentage of positivity violating records using our density partitioning algorithm. 

Sparseness of a slice indicates low probability of a record to have features that abide the slice's rule. Therefore, when looking at sparse slices conditioned on the treatment feature (in this case, TRAINING), the probability of an unconditioned record in the same slice to receive that specific treatment assignment is low. Hence, sparse slices conditioned by TRAINING are likely to have a high percentage of violating records when removing that condition.

Our density partition algorithm supports exactly this type of conditioning; the algorithm can be told to build a separate regression tree on the observations with each level of a categorical conditioning variable. When given a conditioning feature, the density partition algorithm first splits the data on each value of the conditioning feature, and then calculates the rest of the tree, and so each slice includes a condition on this specific feature as well as other conditions on other features.

Therefore, to find slices with high violation percentage, we can condition our slices on TRAINING.  For a given baseline (non-treatment) feature slice, such as $\{\textrm{HISPANIC} \in\{\textrm{True}\}\}$, if the density differs substantially when additionally conditioned on different treatment levels, this may indicate violation of positivity.  The intuition is that low density (sparsity) should correspond roughly with a low number of observations in a given region, and that positivity is violated if a given feature group has very low representation in some values of the treatment, and so outcome effect for this group cannot be estimated well across different treatments.  In this case, with the treatment feature being binary, we need only check the opposite TRAINING value found.
For instance, if conditioning identified the slice $\{\textrm{TRAINING} \in\{\textrm{False}\}\}\:\&\:\{\textrm{HISPANIC} \in\{\textrm{True}\}\}$ as being particularly sparse, if the `opposite' treatment slice 
$\{\textrm{TRAINING} \in\{\textrm{True}\}\}\:\&$ $\{\textrm{HISPANIC} \in\{\textrm{True}\}\}$ is relatively dense, then perhaps the subset $ \{\textrm{HISPANIC} \in\{\textrm{True}\}\}$ (for all treatment groups) should be omitted as it may violate positivity.

Filtering after checking the opposite treatment is needed because the sparse slices we choose may remain sparse after removing the treatment condition. In that case, the probability of a record to be in the slice is low for a reason other than the treatment.  For example, the slice defined by $\{\textrm{BLACK} \in\{\textrm{True}\}\} \:\&\: \{\textrm{HISPANIC} \in\{\textrm{True}\}\} \:\&\: \{\textrm{TRAINING} \in\{\textrm{True}\}\}$ is sparse because there are no black Hispanics in the dataset, and not because of the TRAINING condition. This does not indicate a low probability for a record in the slice to receive this treatment assignment, but just a low probability in general for the baseline feature combination, and so the slice $\{\textrm{BLACK} \in\{\textrm{True}\}\} \:\&\: \{\textrm{HISPANIC} \in\{\textrm{True}\}\}$ doesn't necessarily have a high violation percentage.  Because of this, the relevant slices to omit for causal inference are those that specifically have very different densities depending on which treatment level they are conditioned on.


Note, however, that while having similarities in their objective targets, these two algorithms present different approaches and so do not completely correspond with one another.
In addition, we may receive different results because the density partition doesn't target the causal inference task directly.

\subsubsection{Results}
When running our method on the Lalonde dataset, we get many slices with relatively high violation percentage.  For example, we get the slice defined by $\{32<\:\textrm{AGE}\:<55\}\:\&\: \{\textrm{HISPANIC} \in\{\textrm{True}\}\} \:\&\: \{\textrm{NO-DEGREE} \in\{\textrm{True}\}\} \:\&\: \{\textrm{TRAINING} \in\{\textrm{True}\}\}$, which after removing the TRAINING condition has a support of 442 records, and violation proportion of 0.86.
Our approach in the density partitioning isn't to maximize the number of violating records, but just to divide the space by the density. 
Because of this, the slices we receive may be smaller and so we include slices with a higher number of records, sometimes at the expense of the violating percentage. 
\begin{table}[ht]
\centering
\small
\begin{tabular}{ |c c c|c|c|c|} 
\hline
\multicolumn{3}{|c|}{Feature Rules} & \multicolumn{2}{c|}{Treated} & \multicolumn{1}{c|}{Size}\\
\cline{4-5}

 \multicolumn{3}{|c|}{} &True & False &\\
  \hline
  $\textrm{HISPANIC}\in\{\textrm{True}\}$   & $\textrm{NO-DEGREE}\in\{\textrm{True}\}$ &  32 $<$ AGE $<$ 55 &0 &442 &442 \\ 
  \hline
  $\textrm{BLACK}\in\{\textrm{False}\}$    & $\textrm{NO-DEGREE}\in\{\textrm{True}\}$  & $\textrm{HISPANIC}\in\{\textrm{False}\}$     &9 &12,831 &12,840\\ 
  \hline
  $\textrm{HISPANIC}\in\{\textrm{True}\}$    & $\textrm{NO-DEGREE}\in\{\textrm{True}\}$ &  29 $<$ AGE $<$ 31 &2 &59 &61\\
  \hline
\end{tabular}
\caption{\label{tab:density_violating_slices}List of filtered slices.}
\end{table}

Table~\ref{tab:density_violating_slices} describes the violating slices found. 
Though the slices returned by the density partition are different from the slices found in \cite{SKD2020}, they have similarities. 
While some of the slices received by the density partition have a low percentage of record intersection with the decision tree slices, some have up to 70\% overlap.
 Notice that there is also some correspondence between the slice rules. While we receive condition on age such as $\{32<\:\textrm{AGE}\:<55\}$, the decision tree slice receives similar conditions on age (e.g. $\{\textrm{AGE}\:>39\}$). Curiously, while we receive $\{\textrm{HISPANIC} \in\{\textrm{True}\}\}$   conditions in all slices, the decision tree slices usually include a $\{\textrm{BLACK} \in\{\textrm{False}\}\}$ condition. At first glance these seem like completely different conditions, but after a deeper look we can see there is indeed a relation between the two; They both target ``non-black" people. So while there is a difference in the condition (likely caused by the different approaches of the algorithms), there is resemblance in their targeted records, with non-empty intersection. Another interesting aspect is that while the decision tree slices conditions on the numeric RE75 and RE74 features, our slices don't condition on them at all and instead condition on other categorical features such as NO-DEGREE. This difference may be due to aspects of the density approach, for example the difference between the influence of numeric features and of categorical features on the Gower distance calculation.
\\

After uniting slices, we remove them from the dataset.

\begin{table}[ht]
\centering
\begin{tabular}{|l|c|c|c|c|c|c|}
\hline
\multicolumn{1}{|c|}{} & \multicolumn{3}{c|}{violating} & \multicolumn{3}{c|}{treated}\\
\cline{2-7}
\multicolumn{1}{|c|}{to remove} & False & True &violating portion  & False & True &treated portion \\
\hline
False &5,095  &3,668 &0.41     &8,578 &174 &0.94 \\
True  &3,307  &10,036 &\textbf{0.75}    &13,332 &11 &\textbf{0.001} \\
\hline
\end{tabular}
\caption{\label{tab:removed_slices_res}Statistics of removed slices.}
\end{table}

Table~\ref{tab:removed_slices_res} illustrates the results of the slices that were combined and removed. The number of records removed adds up to 13,343 ($= 3307 + 10,036$) individuals. Out of these, 75\% ($= 10,036/13,343$) were `violating' and only 11 individuals who received treatment were removed.
After removing the slices, we estimated the effect which rose to $-\$1969$. While this number is still negative, we do see an increase from $-\$2658$, the effect estimated before removing the slices.

\begin{figure}%
    \centering
    \subfloat[\centering]{{\includegraphics[scale=0.4]{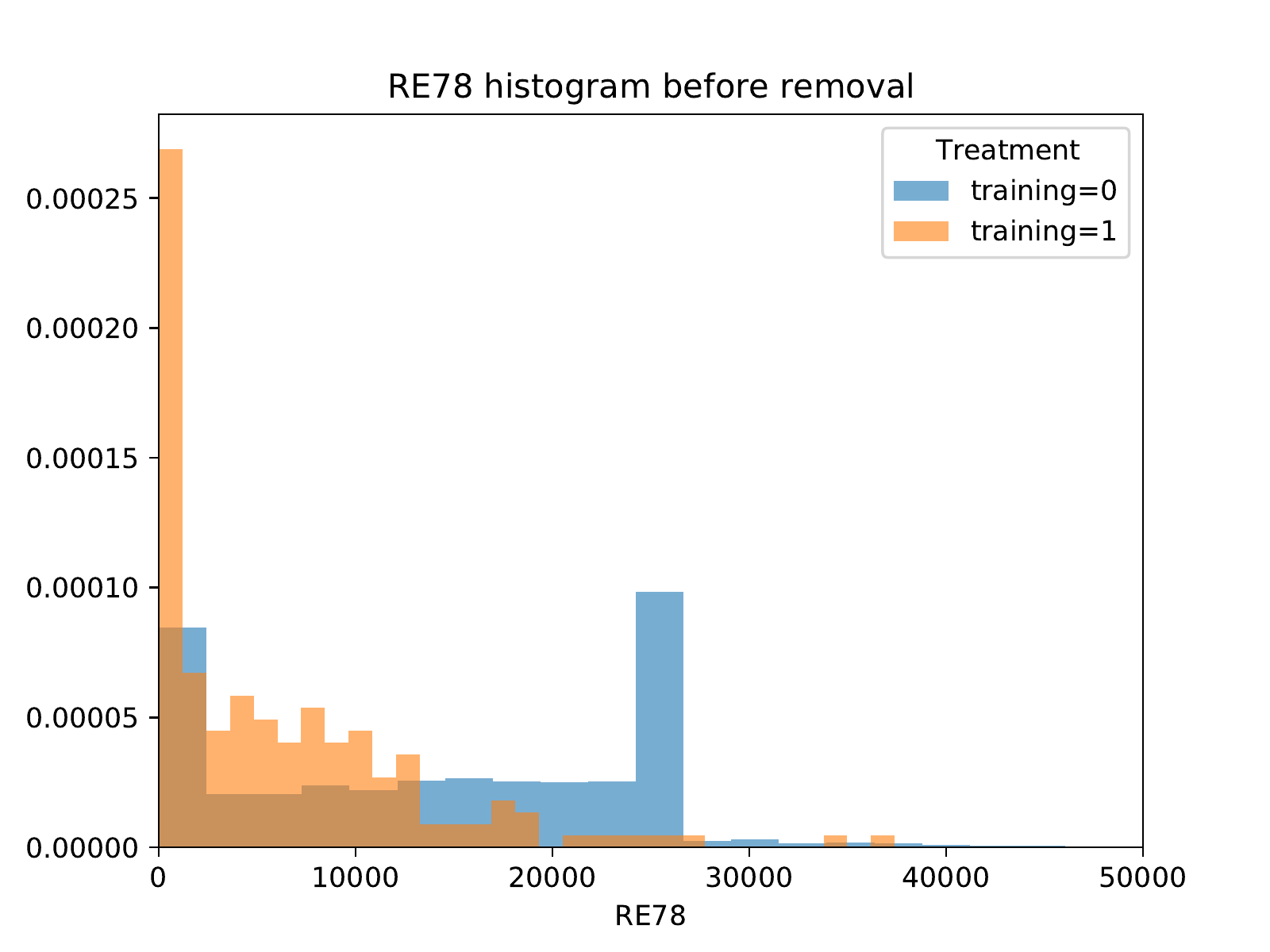}}}%
    \qquad
    \subfloat[\centering]{{\includegraphics[scale=0.4]{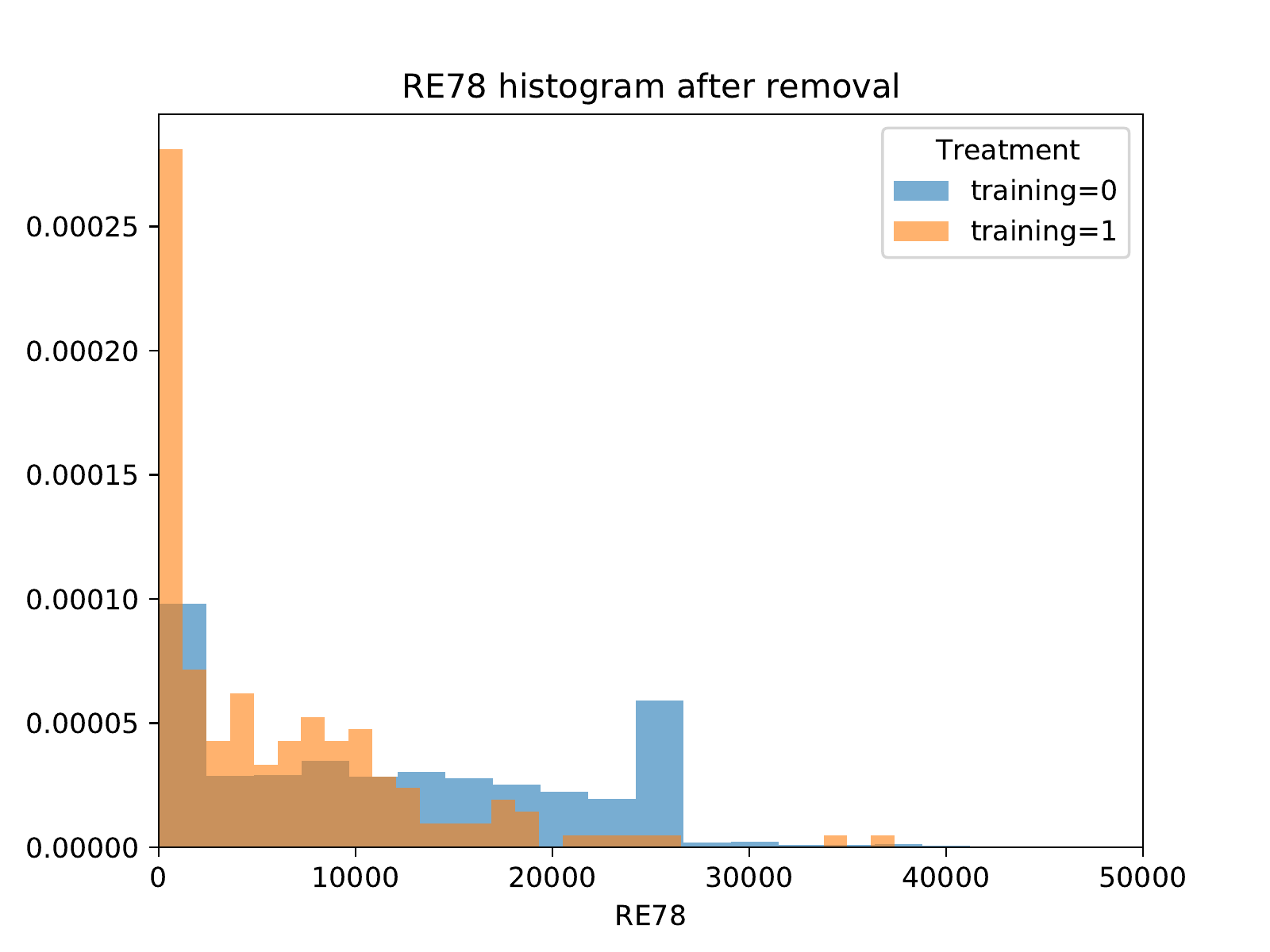} }}%
    \caption{Density histogram of RE78, (a) from before removing the slices from the dataset and (b) from after the removal. The plots show that the graphs of TRAINING=0 and of TRAINING=1 get slightly closer after the removal.}%
    \label{fig:RE78_histogram}%
\end{figure}

In conclusion, though not built for this task, we are able to use the density partition algorithm to find slices that violate the positivity assumption. By removing these slices when performing causal inference, we are able to increase the validity of the analysis. 
These results show great promise in further research on the topic.

\section{Conclusion
\label{sec:conclusion}}

We have presented a method that partitions the observed feature space of a dataset into human-interpretable slices, which allow a user to understand how the observations are distributed in the feature space.  If some areas are dense while large portions of the space are sparse or dense, this is important for data exploration and deployment of ML solutions.  Furthermore, we have illustrated several applications of our technique to related ML tasks, as well as additional analysis metrics that can be used on the resulting partition.  Interested readers are referred to Appendix~\ref{ssec:volume_example} for a full example of the partition calculations.

\itemsep=0pt
{\footnotesize
\printbibliography
}

\newpage
\appendix
\section{Appendix
\label{sec:appendix}}

\subsection{Notation glossary 
\label{ssec:notation}}

\begin{table}[ht]
\centering
\begin{tabular}{l|ll}
  \hline
Notation & Definition & Reference\\
\hline
$p$ & Number of features in dataset $D$ & Section~\ref{sec:problem}\\
$p^*$ & Maximum allowed dimension of a slice $S_i$ & Section~\ref{ssec:feature_domains}\\ 
$F_j$ & $\nth{j}$ dataset feature & Section~\ref{sec:problem}\\
$\textrm{dom}(F_j)$ & Domain of feature $F_j$ & Section~\ref{ssec:feature_domains}\\
$\mathcal{S}$ & Full feature space of dataset $D$ & Section~\ref{ssec:slice_size} \\
$S_i$ & A slice or hyper-rectangle of features, indexed $i$ & Section~\ref{ssec:feature_domains}\\
$s_{i,j}$ & Subset of $\textrm{dom}(F_j)$, indexed $i$ (particularly defining slice $S_i$) & Section~\ref{ssec:feature_domains}\\
$s'_{i,j}$ & A subset of $s_{i,j}$ that is empty & Section~\ref{ssec:empty_space}\\
$|\cdot|$ & Raw (not fractional) length of a subset & Section~\ref{ssec:slice_size}\\
$L(\cdot)$ & Fractional length of a subset $s_{i,j}$ for a given feature & Section~\ref{ssec:slice_size},~\ref{ssec:edge_length}\\
$\textrm{min}_L$ & Minimum fractional length to carve out an empty subset $s_{i,j}$ & Section~\ref{ssec:empty_space}\\
$V(\cdot)$ & Volume of a given slice $S_i$ & Section~\ref{ssec:slice_size},~\ref{ssec:edge_length}\\
\hline
\end{tabular}
\caption{\label{tab:glossary} Glossary of notation.}
\end{table}

\subsection{Edge length and volume calculation
\label{ssec:edge_length}}

As noted in Section~\ref{sec:problem}, a slice $S_i$ is defined as $\bigcap_{j=1}^p s_{i,j}$, where each subset $s_{i,j}\subseteq \textrm{dom}(F_j)$.  Furthermore, the slice volume $V(S_i)=\prod_{j=1}^p L(s_{i,j})$, where $L(s_{i,j})=|s_{i,j}|/|\textrm{dom}(F_j)|$ is the length of $s_{i,j}$ as a fraction of the domain of feature $F_j$.  

Slices, whether or not they contain observations, must have positive volume; therefore we must have $|s_{i,j}|>0,\:\forall j=1,\dots,p$ which define a slice $S_i$. Subsets $s_{i,j}=[a,\:b]$ on integer or ordered categorical features cannot be defined to contain no space.  For ordinal categorical, $s_{i,j}=[a,\:a]$ contains the single level $a$; for integer, $s_{i,j}$ containing the single integer $a'$ is redefined as $s_{i,j}=[a,\:b]=[a'-0.5,\: a'+0.5]$, so that an integer $a'$ contains the space of length 0.5 to left and right between it and the integers $a'\pm 1$.  For a nominal categorical feature $F_j$, if $s_{i,j}=\emptyset$ is the empty set, it will have $|s_{i,j}|=0$.  But such a subset will never be considered, because regression tree nodes splitting on $F_j$ will always contain at least one value.  Also, say the tree has been split on a different feature $F_k$; if the observations in both binary nodes (i.e., slices that are not defined on $F_j$) both contain examples of all levels in $\textrm{dom}(F_j)$, there is no empty space for $F_j$, and empty space must have $L(s_{i,j})>0$.  Therefore, $|s_{i,j}|>0$ always for subsets that will be encountered for integer, and ordinal and nominal categorical features $F_j$.

However, we do want to allow subsets like $s_{i,j}=[a,\:a]$ when $F_j$ is real-valued; as above, such an interval will only be used in a slice definition if it contains observations, and cannot be used in defining an empty region $S_i$.  For instance, if the feature $F_j$=INCOME is real-valued (i.e., including cents and not just whole dollars) $s_{i,j}=[\$0.00, \:\$0.00]$ may be nonempty, containing those with zero income; in fact, this may be a particularly dense interval due to the particular significance of the dollar value 0.  Thus, we want $|s_{i,j}|>0$ even though mathematically, $a-a=0$.  Thus, $s_{i,j}=[a.\:a]$ will only be relevant if it contains observations, but the calculation of $|s_{i,j}|$ will allow us in the same way to calculate $|s_{i,j}|$ for $s_{i,j}=[a,\:b],\:a\ne b$, which either do or do not contain observations.

Our procedure is as such: we pre-remove all features of any type containing only a single value, so if $\textrm{dom}(F_j)=[a,\:b]$ for real-valued feature $F_j$, we must have $a\ne b$.  
Let $n_j\geq 2$ and $n_{i,j}$ be the number of unique values observed for $F_j$ overall in $D$ and in the interval $s_{i,j}$, respectively. As will be seen, if $s_{i,j}$ contains observations, $n_{i,j}=1$ if $a=a$, otherwise $n_{i,j}>1$ (at least the two endpoints $a$ and $b$).  Mathematically, for a real-valued variable $F_j$, a point (single observed value $a$), represented by the interval $[a,\:a]$, has length 0, and so does the \textit{sum} of lengths of intervals containing the individual observed values of $F_j$.  However, for the sake of computation, we will instead assign a small positive constant $0 < \epsilon < 1$ (e.g., $\epsilon=0.001$) to represents the `fractional space' taken up by the observed values of $F_j$.  It is also preferable that $\epsilon<\textrm{min}\left(\frac{x_k-x_i}{|\textrm{dom}(F_j)|}\colon\: x_i < x_k\right)$, where $x_i,\:x_k$ are observed values of $F_j$, that is, the smallest fractional distance between two consecutive different observed values when sorted.  This ensures that for consistency, any positive-length interval $[x_i,\:x_k]$ will have its adjusted length $L$ reflect more the value $x_k-x_i$ (raw length) than the number of unique values it contains.

Thus, for intervals $s_{i,j}$ of form $[a,\;b]$, $(a,\:b)$, $(a,\:b]$, or $[a,\:b)$ of a real-valued feature $F_j$, we define \[L(s_{i,j})=(1-\epsilon)\frac{b-a}{|\textrm{dom}(F_j)|} + \epsilon\frac{n_{i,j}}{n_j}\]
$(b-a)/|\textrm{dom}(F_j)|$ is the `raw' fractional length without the adjustment.  Consider, for real-valued feature $F_j$ with $\textrm{dom}(F_j)=[a,\:b]$, any set of $K$ intervals $\{s_{i,j}\}_{i=1}^K$ which fully partition $\textrm{dom}(F_j)$, where $s_{i,j}$ has endpoints $a_{i,j},\:b_{i,j}$.  The first and last intervals $s_{1,j}$ and $s_{K,j}$ must be closed on the left and right, respectively (containing the limits of the domain $a$ and $b$), but the other ones can be open or closed on either end, as appropriate.  The raw fractional lengths of intervals must have $\sum_{i=1}^K \frac{b_{i,j}-a_{i,j}}{b-a}=1$.  Similarly, if $n_{i,j}$ is the number of unique observation values in $s_{i,j}$ (0 if $s_{i,j}$ contains none), and $n_j$ is the total number of unique observed values of feature $F_j$, $\sum_{i=1}^K \frac{n_{i,j}}{n_j}=1$.  This is because $n_j=\sum_{i=1}^K n_{i,j}$ since the intervals $s_{i,j}$ are a partition (don't overlap and cover $\textrm{dom}(F_j)$ completely), and thus any mutual intersection is empty.  Thus, any unique observed value of $F_j$, of which there are $n_j$, appears in \textit{exactly one} of the $s_{i,j}$, and any unique value in any $s_{i,j}$ must be one of the $n_j$ unique observed values of $F_j$.

Thus, we have 
\[\sum_{i=1}^K L(s_{i,j})=(1-\epsilon)\left(\sum_{i=1}^K\frac{b_{i,j}-a_{i,j}}{b-a}\right) + \epsilon\left(\sum_{i=1}^K \frac{n_{i,j}}{n_j}\right)=(1-\epsilon)1 + \epsilon(1)=1\].

The raw lengths of all slices thus sum up to $1-\epsilon$, while the remaining $\epsilon$ area is used to adjust according to the observed values.  The sum of lengths of subsets in any partition of $\textrm{dom}(F_j)$ must sum to 1, while each $L(s_{i,j})>0$ for each $s_{i,j}$ in the partition, which is trivial to show for the other feature types.  This is necessary because $L(s_{i,j})$ must able to be determined while the density partition is occurring (e.g., to decide if empty space is to be carved out) without the other slices in the partition being determined.

Note that for $p=1$ (a single feature $F_1=F_j$), each interval $s_{i,j}=s_{i,1}$ is its own slice $S_i$ (as in Section~\ref{ssec:region_partition}).  Thus, since $\mathcal{S}=\textrm{dom}(F_1)$, the partition $\{s_{i,1}\}$ of $\textrm{dom}(F_1)$ also partitions $\mathcal{S}$; for single dimensions, we have $V(S_i)=V(s_{i,1})=L(s_{i,1})$, and so $\sum_{i=1}^K L(s_{i,j})=\sum_{i=1}^K V(S_i)=1$, as required.

For example, say $\textrm{dom}(F_j)=[10.95,\: 110.95]$, with length 100 and $n_j=500$ unique values; if $\epsilon=0.001$, the non-empty interval $s_{i,j}=[12.5, 12.5]$, with $n_{i,j}=1$ unique values, would have length $L(s_{i,j})=0.999\frac{0}{100} + 0.001\frac{1}{500} = 0.000002$, rather than $0$.  If $s_{i,j}=[12.5, 62.5]$ (half the span) contained only 3 out of 500 unique values (meaning the distribution is skewed), its length is $L(s_{i,j})=0.999\frac{50}{100} + 0.0001\frac{3}{500}=0.499506$, rather than $0.5$.  Thus, while the adjusted lengths should differ very slightly from the raw lengths, it allows all relevant intervals to have positive length.

\subsection{Volume calculation example
\label{ssec:volume_example}}

Below, we show an illustration of a fully worked-out example of slice volumes based on a density partition.  The partition is done on the Wine quality dataset (originally from \cite{FLAL1988}, available from \cite{UCI}), using only the $F_1$=FLAVANOIDS (real-valued) and $F_2$=PROLINE (integer-valued) features, retaining only the first 50 observations for the sake of compactness.  This shortened dataset is reproduced in Table~\ref{tab:wine_data}.  The density partition was run with $\textrm{min}_L=0.1$ and minimum slice support of 10 observations.

The feature domains of the features are shown in Table~\ref{tab:wine_domain}.  Note that the domain of PROLINE, which is integer-valued, is the observed range extended by 0.5 on either end.

\begin{table}[ht]
\centering
\begin{tabular}{r|rrrrr}
  \hline
Feature $F_j$ & min & max & domain ($\textrm{dom}(F_j)$) & length ($|\textrm{dom}(F_j)|$) & unique values ($n_j$)\\\hline
FLAVANOIDS & 2.19 & 3.93 & (2.19, 3.93) & 1.74 & 42 \\
PROLINE & 680 & 1680 & [679.5, 1680.5] & 1,001 & 42 \\
\hline
\end{tabular}
\caption{\label{tab:wine_domain}Feature summary statistics and domains for wine dataset in Table~\ref{tab:wine_data}.}
\end{table}

\begin{figure}
    \centering
    \includegraphics[scale=1]{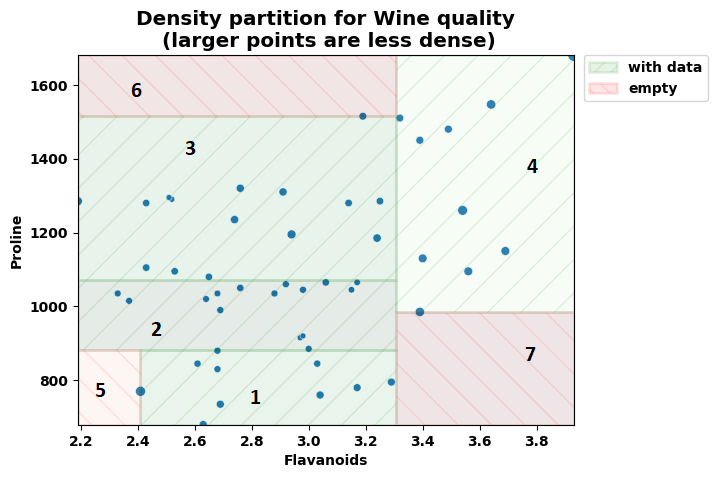}
    \caption{\label{fig:wine_partition}
    Scatterplot of wine quality dataset and density-based partition.  Green slices have observations, and the red ones are empty.  Slice definitions are shown in Table~\ref{tab:wine_slices}.  Slices are labeled according to their ID in the table.
    }
\end{figure}

Based on this dataset, seven slices $\{S_i\}_{i=1}^7$, are determined.  Figure~\ref{fig:wine_partition} plots the scatterplot of observations and overlay of the slice definitions, which are shown in  Table~\ref{tab:wine_slices}; the first four slices (green) have observations, the last three (red) are empty. 

The regional partition proceeded as follows:
\begin{enumerate}
    \item $\mathcal{S}$ split at FLAVANOIDS$\approx3.3$, forming eventual $\{S_1,S_2,S_3,S_5,S_6\}$ to the left and $\{S_4,S_7\}$ to the right.
    \item Empty region $S_7$ was then carved out by finding a significant empty area on the PROLINE axis below 984.5, forming $S_4$.
    \item The left side was split on PROLINE=882.5, forming $\{S_1,S_5\}$ below and $\{S_2,S_3,S_6\}$ above.
    \item On the bottom, $S_5$ is carved out by finding empty space below FLAVANOIDS=2.41, thus forming $S_1$.
    \item On the top, a split was made at PROLINE=1072.5 to form $\{S_2\}$ alone below and $\{S_3,S_6\}$ above. 
    \item $S_2$ was not trimmed since no significant space was found on the edges of the points.
    \item On the top, The empty $S_6$ was carved out, containing the area above PROLINE=1515.5, also forming $S_3$.
\end{enumerate}

In each case here, due to the relative mutual proximity of the observations, all carving out of empty regions was done by the heuristic of finding empty space on any feature after a regression tree split was made, rather than widening the empty space found on the feature used to split a node in the tree (see Section~\ref{ssec:empty_space}).

When slices are defined on the integer-valued PROLINE feature, the bounds are again extended by 0.5 on either end from the range of observed values, to aid in the calculations in Table~\ref{tab:wine_slices_edge_lengths}.  Furthermore, for slices that are empty and defined on the real-valued feature (FLAVANOIDS), the edge will be an interval that is open on at least one of the ends, since it excludes an observed data value that defines the boundary of the neighboring non-empty slice.

\begin{table}[ht]
\renewcommand{\arraystretch}{2}

\centering
\begin{tabular}{r|lr}
\hline
Slice ID $i$ & Slice definition $S_i=s_{i,1}\bigcap s_{i,2}$& Support\\
\hline
1 &  \makecell[l]{$\{2.41 \leq\:\textrm{FLAVANOIDS}\: \leq 3.3049999475479126\}\:\&$ \\ $\{679.5 <\:\textrm{PROLINE}\:< 882.5\}$} & 10
 \\
2 & \makecell[l]{$\{2.19 \leq\:\textrm{FLAVANOIDS}\: \leq 3.3049999475479126\}\:\&$ \\ $\{882.5 <\:\textrm{PROLINE}\:< 1072.5\}$} & 15\\
3 &  \makecell[l]{$\{2.19 \leq\:\textrm{FLAVANOIDS}\: \leq 3.3049999475479126\}\:\&$ \\ $\{1072.5 <\:\textrm{PROLINE}\:< 1515.5\}$} & 15 \\
4 &  \makecell[l]{$\{3.3049999475479126 \leq\:\textrm{FLAVANOIDS}\: \leq 3.93\}\:\&$ \\ $\{984.5 <\:\textrm{PROLINE}\:< 1680.5\}$} & 10 \\
5 &  \makecell[l]{$\{2.19 <\:\textrm{FLAVANOIDS}\: < 2.41\}\:\&$ \\ $\{679.5 <\:\textrm{PROLINE}\:< 882.5\}$} & 0 \\
6 &  \makecell[l]{ $\{2.19 <\:\textrm{FLAVANOIDS}\: < 3.3049999475479126\}\:\&$ \\ $\{1515.5 <\:\textrm{PROLINE}\:< 1680.5\}$} & 0 \\
7 &  \makecell[l]{$\{3.3049999475479126 <\:\textrm{FLAVANOIDS}\: < 3.93\}\:\&$ \\ $\{679.5 <\:\textrm{PROLINE}\:< 984.5\}$} & 0 \\
\hline
\end{tabular}
\caption{\label{tab:wine_slices} Slice definitions found on wine quality dataset.}
\end{table}

Table~\ref{tab:wine_slices_edge_lengths} shows the raw edge lengths (see Appendix~\ref{ssec:edge_length}, Section~\ref{ssec:slice_size}) for each slice in Table~\ref{tab:wine_slices}, taking the difference between the bounds on each feature in the slice, divided by the domain length (Table~\ref{tab:wine_domain}).  For the feature $F_2$=PROLINE, which is integer-valued, the raw length is the same as the final length $L(s_{i,2})$ for this feature.  Since FLAVANOIDS is real-valued, an additional adjustment is shown in Table~\ref{tab:wine_edge_adjustment}, with $\epsilon=0.001$, as in Appendix~\ref{ssec:edge_length}.

\begin{table}[ht]
\centering
\renewcommand{\arraystretch}{1.5}
\begin{tabular}{r|c|c}

\hline
Slice ID $i$ & FLAVANOIDS ($F_1$) $|s_{i,1}|$ & PROLINE ($F_2$) $|s_{i,2}|$\\
\hline
1 &  $\frac{3.3049999475479126 - 2.41}{1.74}=0.5143677859470761$ & $\frac{882.5 - 679.5}{1001}=0.20279720279720279$\\
2 &  $\frac{3.3049999475479126 - 2.19}{1.74}=0.6408045675562717$ & $\frac{1072.5 - 882.5}{1001}=0.18981018981018982$\\
3 &  $\frac{3.3049999475479126 - 2.19}{1.74}=0.6408045675562717$ &$\frac{1515.5 - 1072.5}{1001}=0.44255744255744256$\\
4 &  $\frac{3.93 - 3.3049999475479126}{1.74}=0.3591954324437285$ &$\frac{1680.5 - 984.5}{1001}=0.6953046953046953$\\
5 &  $\frac{2.41 - 2.19}{1.74}=0.12643678160919553$ & $\frac{882.5 - 679.5}{1001}=0.20279720279720279$\\
6 &  $\frac{3.3049999475479126 - 2.19}{1.74}=0.6408045675562717$ &$\frac{1680.5 - 1515.5}{1001}=0.16483516483516483$ \\
7 &  $\frac{3.93 - 3.3049999475479126}{1.74}=0.3591954324437285$ &$\frac{984.5 - 679.5}{1001}=
0.3046953046953047$ \\
\hline
\end{tabular}
\caption{\label{tab:wine_slices_edge_lengths} Raw edge lengths for each slice in Table~\ref{tab:wine_slices}}
\end{table}

Table~\ref{tab:wine_edge_adjustment} calculates $L(s_{i,1})$, the adjusted length of the slice edge length for the $F_1$ FLAVANOIDS feature, with a slight correction of the raw lengths in Table~\ref{tab:wine_slices_edge_lengths}, using the number of unique values of FLAVANOIDS ($n_{i,1},\:i=1,\dots,7$) in each slice, as shown in Appendix~\ref{ssec:edge_length}.  The slice volume for slice $S_i$ is then $V(S_i)=s_{i,1}\times s_{i,2}$, where $s_{i,2}$ is the raw length of the PROLINE edge from Table~\ref{tab:wine_slices_edge_lengths}. These volumes sum within an error of less than 0.04\% of 1.0, the volume of the full feature space $\mathcal{S}$ (see Section~\ref{ssec:slice_size}); this is due to numerical rounding errors from calculations on floating-point values.

\begin{table}[ht]
\renewcommand{\arraystretch}{1.5}
\centering
\begin{tabular}{r|c |c| c |} 
\hline
\multicolumn{1}{c|}{Slice ID $i$} & \multicolumn{2}{|c|}{FLAVANOIDS ($F_1$)} &  \multicolumn{1}{|c|}{VOLUME $V(S_i)$} \\
\cline{2-3}
\cline{4-4}
\multicolumn{1}{c|}{}

& $n_{i,1}$ & adjusted length $L(s_{i,1})$ & $L(s_{i,1})\times L(s_{i,2})$\\
\hline
1 & 9 & $0.999(0.5143677859470761) + 0.001(\frac{9}{42})=0.5140677038754148$ & 0.10425149\\
2 & 14 & $0.999(0.6408045675562717) + 0.001(\frac{14}{42})=0.6404970963220488$ & 0.12157288\\
3 & 14 & $0.999(0.6408045675562717) + 0.001(\frac{14}{42})=0.6404970963220488$ & 0.28345676\\
4 & 9 & $0.999(0.2442529037080962) + 0.001(\frac{9}{42})=0.2442229365186738$ & 0.24975027\\
5 & 0 & $0.999(0.20279720279720279) + 0.001(\frac{0}{42})=0.20259440559440559$ & 0.02564103\\
6 & 0 & $0.999(0.6408045675562717) + 0.001(\frac{0}{42})=0.6401637629887155$ & 0.10562713\\
7 & 0 & $0.999( 0.3591954324437285) + 0.001(\frac{0}{42})=0.35883623701128475$ & 0.10933572\\
\hline
\end{tabular}
\caption{\label{tab:wine_edge_adjustment} Adjusted edge length for FLAVANOIDS, and slice volume.  Here, $\epsilon=0.001$.}
\end{table}

\begin{table}[ht]
\centering
\begin{tabular}{r|rr|r}
\hline
{} &  FLAVANOIDS ($F_1$) &  PROLINE ($F_2$) &  Slice ID $i$ \\
\hline
1  &        3.06 &     1065 &      2 \\
2  &        2.76 &     1050 &      2 \\
3  &        3.24 &     1185 &      3 \\
4  &        3.49 &     1480 &      4 \\
5  &        2.69 &      735 &      1 \\
6  &        3.39 &     1450 &      4 \\
7  &        2.52 &     1290 &      3 \\
8  &        2.51 &     1295 &      3 \\
9  &        2.98 &     1045 &      2 \\
10 &        3.15 &     1045 &      2 \\
11 &        3.32 &     1510 &      4 \\
12 &        2.43 &     1280 &      3 \\
13 &        2.76 &     1320 &      3 \\
14 &        3.69 &     1150 &      4 \\
15 &        3.64 &     1547 &      4 \\
16 &        2.91 &     1310 &      3 \\
17 &        3.14 &     1280 &      3 \\
18 &        3.40 &     1130 &      4 \\
19 &        3.93 &     1680 &      4 \\
20 &        3.03 &      845 &      1 \\
21 &        3.17 &      780 &      1 \\
22 &        2.41 &      770 &      1 \\
23 &        2.88 &     1035 &      2 \\
24 &        2.37 &     1015 &      2 \\
25 &        2.61 &      845 &      1 \\
26 &        2.68 &      830 &      1 \\
27 &        2.94 &     1195 &      3 \\
28 &        2.19 &     1285 &      3 \\
29 &        2.97 &      915 &      2 \\
30 &        2.33 &     1035 &      2 \\
31 &        3.25 &     1285 &      3 \\
32 &        3.19 &     1515 &      3 \\
33 &        2.69 &      990 &      2 \\
34 &        2.74 &     1235 &      3 \\
35 &        2.53 &     1095 &      3 \\
36 &        2.98 &      920 &      2 \\
37 &        2.68 &      880 &      1 \\
38 &        2.43 &     1105 &      3 \\
39 &        2.64 &     1020 &      2 \\
40 &        3.04 &      760 &      1 \\
41 &        3.29 &      795 &      1 \\
42 &        2.68 &     1035 &      2 \\
43 &        3.56 &     1095 &      4 \\
44 &        2.63 &      680 &      1 \\
45 &        3.00 &      885 &      2 \\
46 &        2.65 &     1080 &      3 \\
47 &        3.17 &     1065 &      2 \\
48 &        3.39 &      985 &      4 \\
49 &        2.92 &     1060 &      2 \\
50 &        3.54 &     1260 &      4 \\
\hline
\end{tabular}
\caption{\label{tab:wine_data} First 50 observations of wine quality data, for two selected numeric observations.}
\end{table}

\end{document}